\documentclass{article}

 \usepackage[preprint]{neurips_2025}

\usepackage[utf8]{inputenc} % allow utf-8 input
\usepackage[T1]{fontenc}    % use 8-bit T1 fonts
\usepackage{hyperref}       % hyperlinks
\usepackage{url}            % simple URL typesetting
\usepackage{booktabs}       % professional-quality tables
\usepackage{amsfonts}       % blackboard math symbols
\usepackage{nicefrac}       % compact symbols for 1/2, etc.
\usepackage{microtype}      % microtypography
\usepackage{xcolor}         % colors
\usepackage{booktabs} % For professional table rules
\usepackage{array} % For better column formatting
\usepackage{siunitx} % For consistent number formatting
\usepackage{caption} % For better caption control
\usepackage{booktabs}
\usepackage{hyperref}
\usepackage{longtable}
\usepackage{array}
\usepackage{graphicx} % 导入 graphicx 宏包，用于处理图片
\usepackage{float} % 导入 float 宏包，用于更灵活地控制图片位置
\usepackage{ragged2e}
\usepackage{subfig}
\usepackage{CJKutf8}
\makeatletter
\def\UrlAlphabet{%
      \do\a\do\b\do\c\do\d\do\e\do\f\do\g\do\h\do\i\do\j%
      \do\k\do\l\do\m\do\n\do\o\do\p\do\q\do\r\do\s\do\t%
      \do\u\do\v\do\w\do\x\do\y\do\z\do\A\do\B\do\C\do\D%
      \do\E\do\F\do\G\do\H\do\I\do\J\do\K\do\L\do\M\do\N%
      \do\O\do\P\do\Q\do\R\do\S\do\T\do\U\do\V\do\W\do\X%
      \do\Y\do\Z}
\def\UrlDigits{\do\1\do\2\do\3\do\4\do\5\do\6\do\7\do\8\do\9\do\0}
\g@addto@macro{\UrlBreaks}{\UrlOrds}
\g@addto@macro{\UrlBreaks}{\UrlAlphabet}
\g@addto@macro{\UrlBreaks}{\UrlDigits}
\makeatother

\title{CCI4.0: A Bilingual Pretraining Dataset for Enhancing Reasoning in Large Language Models}

\author{
  Guang Liu\thanks{Corresponding Author, liuguang@baai.ac.cn} \quad Liangdong Wang \quad Jijie Li \\
  \textbf{Yang Yu} \quad \textbf{Yao Xu} \quad \textbf{Jiabei Chen} \quad \textbf{Yu Bai} \quad \textbf{Feng Liao} \quad \textbf{Yonghua Lin}\\
  Data Research Team \\ 
  Beijing Academy of Artificial Intelligence \\
}

\begin{document}

\begin{CJK}{UTF8}{gbsn}
\maketitle

\begin{abstract}
We introduce CCI4.0, a large-scale bilingual pre-training dataset engineered for superior data quality and diverse human-like reasoning trajectory. CCI4.0 occupies roughly $35$ TB of disk space and comprises two sub-datasets: CCI4.0-M2-Base and CCI4.0-M2-CoT. CCI4.0-M2-Base combines a $5.2$ TB carefully curated Chinese web corpus, a $22.5$ TB English subset from Nemotron-CC, and diverse sources from math, wiki, arxiv, and code. Although these data are mostly sourced from well-processed datasets, the quality standards of various domains are dynamic and require extensive expert experience and labor to process. So, we propose a novel pipeline justifying data quality mainly based on models through two-stage deduplication, multiclassifier quality scoring, and domain-aware fluency filtering. We extract $4.5$ billion pieces of CoT(Chain-of-Thought) templates, named CCI4.0-M2-CoT. Differing from the distillation of CoT from larger models, our proposed staged CoT extraction exemplifies diverse reasoning patterns and significantly decreases the possibility of hallucination. Empirical evaluations demonstrate that LLMs pre-trained in CCI4.0 benefit from cleaner, more reliable training signals, yielding consistent improvements in downstream tasks, especially in math and code reflection tasks. Our results underscore the critical role of rigorous data curation and human thinking templates in advancing LLM performance, shedding some light on automatically processing pretraining corpora. 
\end{abstract}

\bibliographystyle{plain}

%\iffalse written by Yang Feng
\section{Introduction} \label{Introduction}

In recent years, large language models (LLMs) have achieved remarkable success across a broad spectrum of natural language processing tasks, including text generation, translation, and sentiment analysis. A critical factor underpinning these advances is the availability and quality of large-scale pre-training data \cite{li2024datacomplmsearchgenerationtraining,huggingface2024fineweb,gao2020pile}. Pre-training data not only shapes the linguistic capabilities of LLMs but also plays a central role in determining their generalization and reasoning abilities across a wide range of downstream applications.

Despite growing efforts to curate and release open-source datasets for language model training \cite{su2024nemotroncctransformingcommoncrawl,soldaini2024dolma,he2023wanjuancomprehensivemultimodaldataset}, there remains a significant gap in the availability of high-quality and diverse large-scale corpora. Most existing resources are limited in either linguistic diversity or domain coverage, constraining the models’ ability to generalize beyond narrow contexts. Furthermore, although real-world data is often prioritized for its authenticity, synthetic data has emerged as an important complement, particularly in fostering reasoning skills. Nevertheless, high-quality synthetic pre-training datasets—especially those that explicitly incorporate structured reasoning—are still notably lacking.

To address the aforementioned gaps and promote the advancement of data-centric development in large language models, we introduce CCI4.0-M2-Base: a high-quality and diverse bilingual corpus in Chinese and English. In particular, CCI4.0-M2-Base includes a large-scale, bilingual pretraining dataset (35T tokens) combining a Chinese corpus and Nemotron-CC’s English data, with a meticulously-designed pipeline to yield a high-quality dataset to enhance LLM general and reasoning capabilities. Moreover, to enhance the diversity of the corpus, we additionally incorporate high-quality source data from a wide range of domains, including web pages, code, mathematics, academic papers, and encyclopedias.  

Furthermore, considering that the reasoning capabilities of LLMs are primarily developed during the pretraining phase \cite{gandhi2025cognitive,yue2025does,reflection2025}, we provide CCI4.0-M2-CoT, which integrates 4.5 billion human thinking templates synthesized from high-quality samples using advanced techniques. While effective for general language understanding, traditional pretraining datasets often lack the specialized content needed to foster advanced reasoning skills, such as explicit representations of human thought processes or logical reasoning traces. To address this gap, we introduce CCI4.0-M2-CoT, whose templates are crafted to embed diverse reasoning patterns, strengthening the foundational reasoning abilities of LLMs, and decreasing the possibility of hallucination.

Experimental results validate the efficacy of CCI4.0, demonstrating substantial performance improvements on knowledge-based and reasoning-intensive benchmarks such as MMLU and ARC-Challenge, with notable gains in commonsense reasoning and mathematical problem-solving. These findings underscore the critical role of high-quality, diverse and reasoning-focused pretraining data in advancing LLMs’ ability to tackle complex, multi-step reasoning tasks. By addressing the limitations of existing datasets, CCI4.0 sets a new benchmark for pretraining and paves the way for the development of more capable and versatile language models.

This paper makes the following core contributions:
\begin{itemize}
    \item Introduction of CCI4.0-M2-Base: A large-scale, bilingual pretraining dataset (35T tokens) combining a Chinese corpus, Nemotron-CC’s English data and corpora sourced from diverse domains, designed to enhance LLM general and reasoning capabilities.
    \item Incorporation of CCI4.0-M2-CoT: Integration of Diverse Reasoning Templates including 4.5 billion synthesized human thinking templates, embedding diverse reasoning patterns to bolster logical and commonsense reasoning and decrease the hallucination.
    \item Advanced Data Processing Pipeline: A comprehensive methodology including deduplication, multi-classifier quality scoring, fluency filtering, CoT synthesis, and privacy/toxicity handling, ensuring high-quality and diverse data curation.
    \item Empirical Validation: Demonstration of significant performance gains on benchmarks like MMLU and ARC-Challenge, particularly in mathematical problem-solving and commonsense reasoning, outperforming baseline datasets such as Nemotron-CC-HQ and CCI3-HQ.
\end{itemize}

\begin{table}[ht]
\centering
\caption{Dataset Comparison. Column abbreviations: Size (Open Source Size), Multi-Src (Multi-Source), Multi-Cls (Multiple Classifiers), Multi-Lang (Multilingual), CoT-Syn (CoT Synthesis).}
\label{tab:dataset_comparison}
\resizebox{0.8\textwidth}{!}{
\begin{tabular}{cccccc}
% {>{\raggedright\arraybackslash}p{2.5cm}>{\raggedright\arraybackslash}p{2cm}cccc}
\toprule
\textbf{Dataset} & \textbf{Size(TB)} & \textbf{Multi-Src} & \textbf{Multi-Cls} & \textbf{Multi-Lang} & \textbf{CoT-Syn} \\
\midrule
Pile & 0.8 & $\checkmark$ & $\times$ & $\checkmark$ & $\times$ \\
Dolma & 11 & $\checkmark$ & $\checkmark$ & $\times$ & $\times$ \\
RefindeWeb & 0.6 & $\times$ & $\times$ & $\checkmark$ & $\times$ \\
Redpajama (V2) & $\sim$120 & $\times$ & $\checkmark$ & $\checkmark$ & $\times$ \\
FineWeb2 & 17.7 & $\times$ & $\times$ & $\checkmark$ & $\times$ \\
Wanjuan1.0 & 0.19 & $\times$ & $\checkmark$ & $\checkmark$ & $\times$ \\
FineWeb & 44 & $\times$ & $\times$ & $\times$ & $\times$ \\
Nemotron-CC & 22 & $\times$ & $\checkmark$ & $\times$ & $\times$ \\
CCI3.0 & 1 & $\times$ & $\checkmark$ & $\times$ & $\times$ \\
CCI4.0 & 35 & $\checkmark$ & $\checkmark$ & $\checkmark$ & $\checkmark$ \\
\bottomrule
\end{tabular}
}
\end{table}
\section{Related Works}
To contextualize the development of CCI4.0, we compare it with existing large-scale pretraining datasets, as summarized in Table \ref{tab:dataset_comparison}. The Pile (0.8T tokens) and Dolma (11T tokens) leverage multi-source data and multiple classifiers for quality assurance, supporting multilingual content but lacking Chain-of-Thought (CoT) synthesis, which limits their focus on reasoning enhancement. Similarly, datasets like Redpajama (V2) (~120T tokens) and FineWeb2 (17.7T tokens) offer substantial scale and multilingual support, yet they do not incorporate CoT synthesis or multi-source diversity to the extent of CCI4.0. Nemotron-CC (10.4T tokens), a key component of CCI4.0’s English corpus, employs multiple classifiers but is monolingual and lacks CoT integration. In contrast, CCI4.0 (35T tokens) uniquely combines multi-source data, multilingual support (English and Chinese), multiple quality classifiers, and CoT synthesis, incorporating 4.5 billion human thinking templates to explicitly target reasoning capabilities. This comprehensive approach distinguishes CCI4.0 from prior datasets, addressing gaps in reasoning-focused pretraining data and setting a new benchmark for developing LLMs with enhanced logical and commonsense reasoning abilities.

\section{Method} \label{method}

\begin{figure}[H] % H 表示将图片精确地放置在此处 (需要 float 宏包)
    % 其他可选的位置参数:
    % h: here - 尽量放在当前位置
    % t: top - 放在页面顶部
    % b: bottom - 放在页面底部
    % p: page of floats - 放在一个专门的浮动页面
    % !: 忽略 LaTeX 的一些内部参数，强制执行位置参数

    \centering % 图片居中显示
    \includegraphics[width=1.0\textwidth]{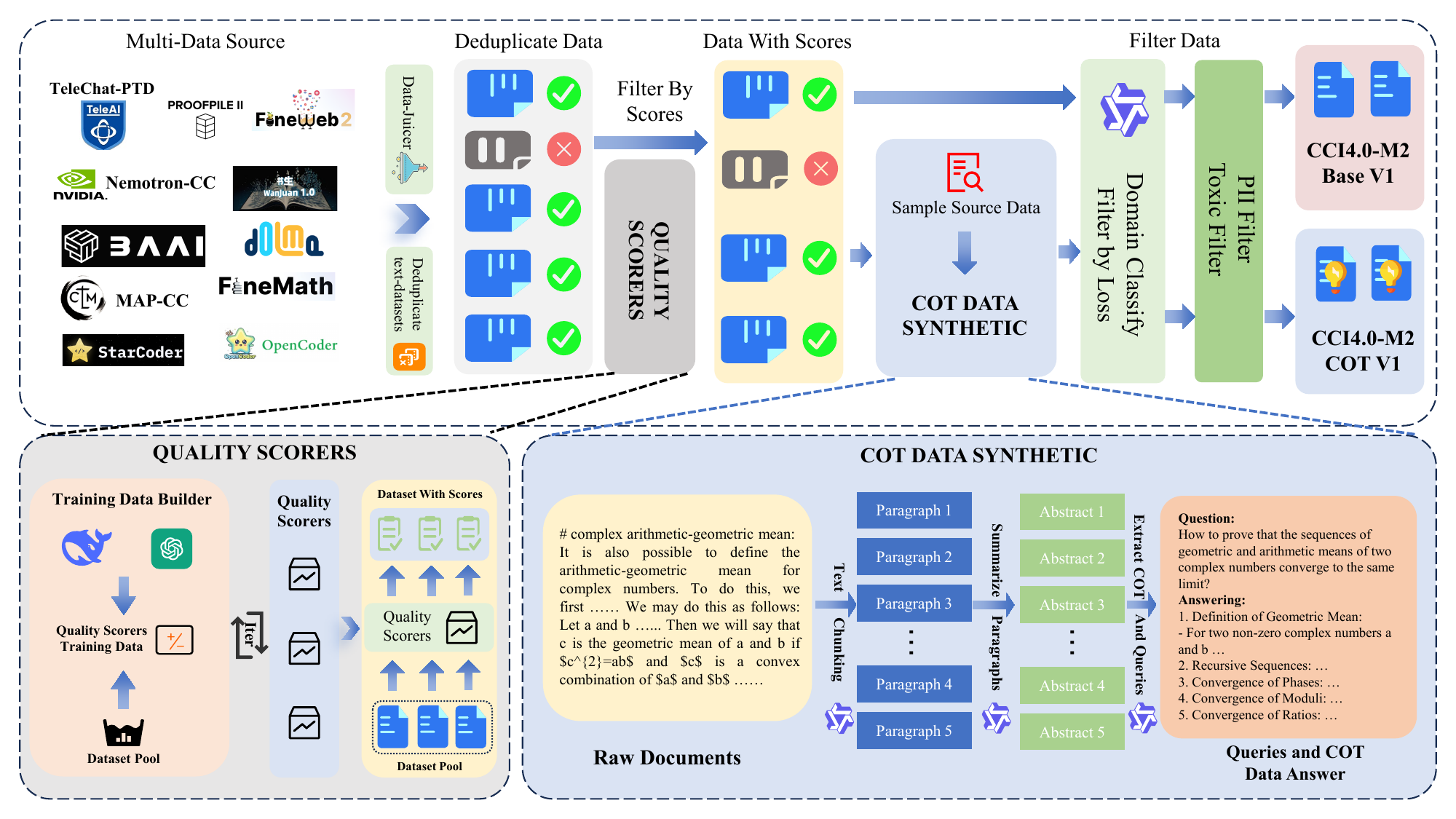} % 插入图片
    % width=0.8\textwidth 表示图片宽度为文本宽度的 80%
    % 你也可以使用其他单位，例如:
    % height=5cm 表示图片高度为 5 厘米
    % scale=0.5 表示图片缩放为原来的 50%
    % angle=45 表示图片旋转 45 度
    % {example-image} 是图片的文件名 (不需要写扩展名，LaTeX 会自动查找支持的格式，如 .png, .jpg, .pdf, .eps)
    % 确保图片文件与你的 .tex 文件在同一目录下，或者提供正确的相对/绝对路径

    \caption{The figure shows the overall processing pipeline of our dataset. Our main processing flow includes deduplication, scoring the data, and some other filtering process to get our CCI4.0-M2-Base V1 dataset. Notably, we incorporated COT synthetic data by leveraging large language models to perform chunking, summarization, and extraction operations on the original documents sampled from our dataset pool to get the CCI4.0-M2-COT V1 dataset.} % 图片的标题
    \label{fig:framework} % 图片的标签，用于在文本中引用，例如 \ref{fig:example}
\end{figure}
As shown in Figure~\ref{fig:framework}, our data processing pipeline is meticulously designed to yield a high-quality, diverse, and robust dataset, comprising five principal stages: \textbf{Deduplication}, \textbf{Multi-classifier Quality Scoring}, \textbf{Fluency Filtering}, \textbf{Data Synthesis (including CoT synthesis)}, and \textbf{comprehensive Privacy and Toxicity Handling}. The initial Deduplication phase is critical for removing redundancy, operating at both a global document level and a finer-grained string level. Following this, the Multi-classifier Quality Scoring stage evaluates data integrity and relevance across various dimensions. This is achieved by automatically constructing evaluation samples using large language models, training specialized small-scale quality classifiers on these samples, and then integrating their scores to assign a comprehensive quality tier to each data point. To address linguistic quality, particularly the prevalence of overly short or syntactically challenging samples, a domain-specific Fluency Filtering step is implemented. Recognizing the significant variability in fluency characteristics across different data domains, this filtering process is applied independently to each domain to effectively remove samples with notably poor linguistic flow. Building upon the foundation of high-quality data identified through the previous stages, a Data Synthesis process is employed. This involves leveraging the filtered high-quality samples as seeds with large models to generate novel data instances in diverse formats. Specifically, high-quality sources are selected for targeted Chain-of-Thought (CoT) synthesis, focusing on the construction of core questions and detailed instructions. Finally, the pipeline incorporates essential safety and privacy measures: Privacy Handling processes sensitive personal information such as identification numbers and phone numbers, while Toxicity Scoring utilizes a dedicated model to assess and flag potentially harmful content within each sample.

\subsection{Data Collection and Preprocessing}
As shown in Table \ref {tab:dataset_info}, we select multiple sources from English and Chinese. Regarding data sources, the English web corpus was derived from the Nemotron-CC~\cite{nvidia2024nemotron} dataset. This particular source was selected based on both comparative effectiveness evaluations and its significantly larger data volume compared to alternatives. For the Chinese web dataset, our collection process involved consolidating data from some existing open-source Chinese datasets such as~\cite{baai_cci_data,baai_cci2_data,baai_wudaocorpora,chen2023chinesewebtext} and extracting the Chinese components from various multilingual datasets such as~\cite{he2023wanjuancomprehensivemultimodaldataset,penedo2024fineweb-2,penedo2023refinedweb}. Through a thorough analysis of the inter-dependencies and relationships among these potential sources, we strategically filtered and identified over ten datasets that served as the primary contributors to our Chinese web corpus. Furthermore, to enhance the breadth and depth of the training data, we incorporated additional high-performing open-source datasets. These supplementary sources are designed to cover a diverse array of domains, including but not limited to code, mathematics, books, encyclopedias, and academic papers.

Upon conducting manual spot checks on data samples procured from various sources, we identified specific quality inconsistencies, particularly within the Chinese text and code corpora. Consequently, distinct processing methodologies were developed and applied specifically to address these observed issues. For the Chinese dataset, a series of pre-processing operations were undertaken to ensure corpus quality and optimize its suitability for downstream model training. First, to standardize the linguistic and symbolic representation, all text was uniformly converted to Simplified Chinese. Second, to uphold content integrity and compliance with usage norms, a sensitive word filtering mechanism was implemented to automatically detect and remove segments containing inappropriate vocabulary. Furthermore, to mitigate the risk of the model learning overly short or structurally fragmented sentences, a minimum average line length constraint was imposed, retaining only text samples with an average character count of at least 10 per line. Finally, a filter based on the total character count was applied, limiting samples to a range between 100 and 20,000 characters to balance semantic richness with processing efficiency. In parallel, during the processing of the raw code data, we noted the presence of a significant amount of interspersed copyright declarations and related textual information. To construct a high-quality dataset containing solely code content, this non-code material was systematically filtered and removed.

\subsection{Hybrid Deduplication}
Following the initial phase of basic quality filtering, a two-stage deduplication strategy was implemented to further enhance the purity and uniqueness of the dataset. The first stage employed a fuzzy deduplication approach\cite{Broder1997OnTR}, leveraging the fuzzy deduplication operator available within the Data-Juicer framework\cite{Chen2023DataJuicer1}. This method is adept at identifying and eliminating redundant samples by effectively recognizing pairs of texts that are similar in content but not strictly identical. Subsequently, the second stage utilized the deduplicate-text-datasets library\cite{deduplicate-text-datasets}, an open-source tool developed by Google, to perform exact substring deduplication\cite{lee2022deduplicatingtrainingdatamakes}. This process further removes duplicate data based on precise substring matching. Specifically, we configured the parameters with a length-threshold of 800 and min-doc-words of 35. These settings were carefully chosen to ensure that strict comparisons were performed only between samples exceeding a certain threshold of text length and word count, thereby preventing excessive deduplication of shorter texts. These two complementary deduplication methods work in concert to significantly reduce redundancy while concurrently preserving the diversity of the dataset.

\subsection{Quality Classification}

To ensure the high quality of our processed datasets, a multi-faceted quality classification approach was employed, tailored to the characteristics of both English and Chinese corpora. For the English web data, primarily sourced from Nemotron-CC, three independent quality classifiers were utilized to score each document. Based on these scores, samples were allocated into 20 distinct quality bins, with the highest score among the three classifiers designated as the final quality score for each document.

For the Chinese dataset, recognizing the unique linguistic features and the need for domain-specific evaluation, we meticulously designed and trained specialized Chinese quality classifiers during the data construction process. This classification system was conceptually informed by the Nemotron-CC quality classification framework but underwent significant customization and optimization to effectively handle Chinese linguistic nuances and corpus characteristics. The development of the Chinese quality classifiers involved several critical steps\cite{su2024nemotron}. Initially, we devised specific prompts to guide large models in generating the necessary training data for the classifiers. The training sets were generated using two distinct large models, Qwen2.5-72B-Instruct~\cite{qwen2.5} and Deepseek-V3~\cite{deepseekai2025deepseekv3technicalreport}, yielding a total of 460k samples. The test set, comprising 40k samples, was generated by GPT-4o. The initial sample pools contained 1.2M samples for the training set and 116k samples for the test set. In terms of prompt design, Qwen2.5-72B-Instruct utilized a direct scoring prompt in Chinese, while Deepseek-V3 employed a rule-based cumulative scoring prompt in English. This strategic choice aimed to leverage the respective generative strengths of the two models under different prompting styles.

During the training phase, we fine-tuned two independent XLRoberta-based models\cite{Conneau2019UnsupervisedCR} on the respective training sets, resulting in two distinct Chinese quality classifiers. We explored four different learning rates (6e-4, 3e-4, 1e-4, and 6e-5) for each configuration, training a complete model under each setting. Evaluation on the test set revealed that when using each classifier independently, a learning rate of 3e-4 yielded the highest F1 score. Furthermore, we observed a significant improvement in the F1 score when combining the outputs of the two classifiers, indicating the complementary nature of the features captured by the data generated from the two different training sets, thereby enhancing the discriminative power of the combined classification system.

In addition to the XLRoberta-based classifiers, and drawing inspiration from the findings presented in \ref{sec:quality}, where fastText-based filtering demonstrated optimal performance compared to various model-based data filtering strategies, we also trained a fastText\cite{joulin2016bag} classifier specifically for the Chinese corpus scenario. This classifier was designed as a binary classification task to identify high-quality samples. The positive sample pool was initially constructed by collecting multiple Chinese instruction datasets, including COIG-CQIA\cite{Bai2024COIGCQIAQI}, OpenHermes-2.5-zh\cite{OpenHermes2.5-zh}, OpenOrca-Chinese and smoltalk Chinese\cite{yu2025opencsgchinesecorpusseries}. Through multiple iterations, we progressively refined this set by mitigating the influence of training data length distribution, removing irrelevant high-frequency words from predicted positive samples, and adding certain stop words based on word importance features. This iterative process resulted in a final positive sample set of 220k samples. Correspondingly, 220k samples were randomly drawn from the corpus pool to serve as negative samples. These were then used to construct the final training set of 400k samples and a test set of 40k samples, maintaining a 10:1 ratio.

 % The fastText text quality classifier for the Chinese context was ultimately trained using the parameters detailed in the table below:

\subsection{LLM-based Fluency Filtering}
To further refine data quality based on linguistic fluency as \cite{gao2020pile}, we employed a multilingual domain classifier \footnote{\url{https://huggingface.co/nvidia/multilingual-domain-classifier}} to categorize all raw corpus data into distinct domains, resulting in 26 identified sub-domains. Following this domain classification, we computed the Perplexity Loss for all samples within each domain. The analysis of these loss distributions revealed significant variations across different domains. Notably, the 'Games' domain exhibited the highest overall loss values, suggesting that domains such as gaming contain more complex, unpredictable, or specialized linguistic patterns. Conversely, domains like 'Law and Government' and 'Science and Health' showed the lowest average loss, indicating the presence of more established terminology and formal structures. To mitigate the influence of extreme outliers within each domain, we established a filtering criterion based on the calculated loss distributions. Specifically, samples exceeding the $99.5_{th}$ percentile of the loss value within their respective domains were systematically removed, yielding a refined dataset with improved linguistic consistency within each category. Loss and percentiles across domains are illustrated in the Figure \ref{fig:loss} in Appendix.

\subsection{Data synthesis}
Recent studies indicate that the reasoning abilities of large language models (LLMs) primarily originate from the pre-training phase and are subsequently activated during the reinforcement learning phase\cite{gandhi2025cognitive,yue2025does}. Consequently, we endeavor to extract vast quantities of high-quality human thought processes from pre-training corpora to synthesize reasoning data. Specifically, we first curated high-quality source data from diverse domains, including web pages, code, mathematics, academic papers, and encyclopedias. As illustrated in the Figure \ref{fig:framework}, our detailed methodology is as follows:
\begin{itemize}
    \item Semantic Segmentation and Summarization: We utilize Qwen2.5-32B-Instruct\cite{qwen2.5} to perform semantic segmentation on the original texts. This process divides the original documents into semantically independent and non-overlapping segments. To minimize the LLM's output cost, output only contains the start and end markers of each segment. Subsequently, for each segment, we prompt the model to generate a concise summary.
    \item Summarizing Chain-of-Thought and Core Question: Based on the segmented summaries, we have been able to reconstruct the thought process behind the original human-written document. We then employ Qwen2.5-32B-instruct to refine and consolidate these segment summaries, thereby forming a logically coherent chain-of-thought (CoT). Recognizing that recent work highlights questions as a critical element of reasoning data, we finally derive and summarize the core question addressed by the document, based on this CoT.
\end{itemize}
Consequently, each synthesized reasoning data instance is structured as: \{core question, chain-of-thought, original document\}. In total, we have synthesized over \textbf{400 billion (400B)} tokens of reasoning data spanning these diverse domains, including web pages, code, mathematics, academic papers, and encyclopedias.

\section{Experimental Setting} \label{experiment setting}
\subsection{Training Configuration}
    Following the training setup of CCI3-HQ\cite{wang2024cci30hqlargescalechinesedataset}, we adopt the Qwen2-0.5B\cite{qwen2} tokenizer and model architecture, training on a bilingual dataset containing 100 billion tokens. This setup is designed to effectively accommodate both Chinese and English data while preserving consistency across experiments. The training is conducted with a sequence length of 4096, weight decay set to 0.1, and gradient clipping at 1.0. The dataset consists of 25 million samples, trained with a global batch size of 1024. The learning rate follows a cosine decay schedule, starting at 3e-4, decaying to a minimum of 3e-5, with a warmup over the first 2,048,000 samples.
\subsection{Evaluation Metrics}
We used the LightEval\cite{lighteval} library for model evaluation, following the same setup as in FineWeb\cite{huggingface2024fineweb} and CCI3-HQ. All evaluations were conducted in a zero-shot setting. To directly compare the performance across different datasets, we use Average, which refers to the overall average score across all Chinese and English benchmarks. The evaluation metrics include:
\begin{itemize}
\item Chinese benchmarks: CEval\cite{huang2023ceval} and CMMLU\cite{li2024cmmlumeasuringmassivemultitask}.
\item English benchmarks: ARC-C\cite{clark2018thinksolvedquestionanswering}, ARC-E\cite{clark2018thinksolvedquestionanswering}, HellaSwag\cite{zellers2019hellaswagmachinereallyfinish}, WinoGrande\cite{10.5555/3031843.3031909}, MMLU\cite{hendrycks2021measuringmassivemultitasklanguage}, OpenbookQA\cite{banerjee-etal-2019-careful}, PIQA\cite{bisk2019piqareasoningphysicalcommonsense} and SIQA\cite{sap2019socialiqacommonsensereasoningsocial}.
\end{itemize}
\section{Experimental Results}
\subsection{Main Results}
\begin{figure}[htbp] % H 表示将图片精确地放置在此处 (需要 float 宏包)
    % 其他可选的位置参数:
    % h: here - 尽量放在当前位置
    % t: top - 放在页面顶部
    % b: bottom - 放在页面底部
    % p: page of floats - 放在一个专门的浮动页面
    % !: 忽略 LaTeX 的一些内部参数，强制执行位置参数

    \centering % 图片居中显示
   \includegraphics[width=0.75\textwidth]{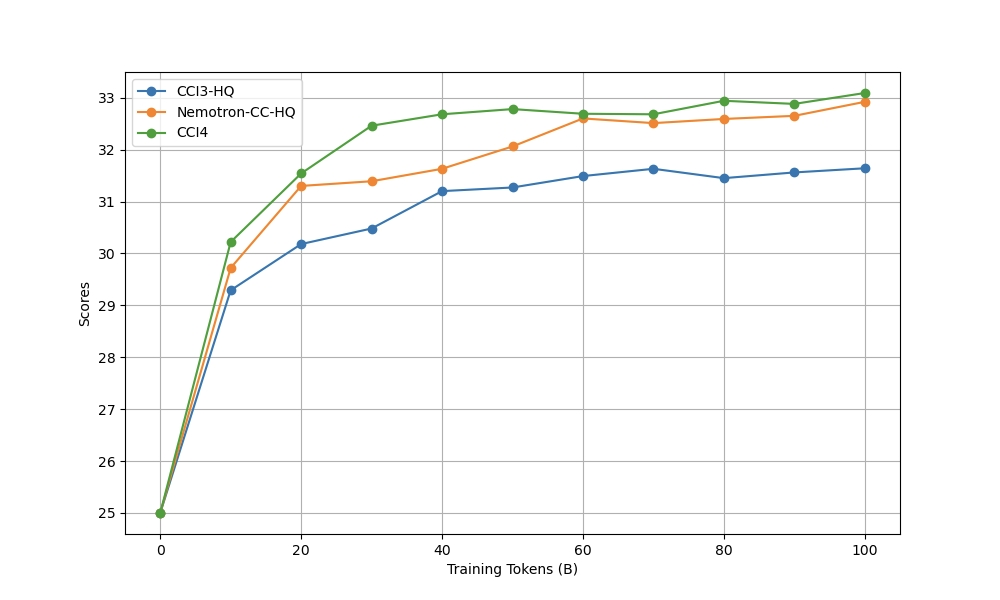} % 插入图片
    
    \caption{Performance Comparison of Different Datasets Across Training Scales.} % 图片的标题
    \label{fig:main} % 图片的标签，用于在文本中引用，例如 \ref{fig:example}
\end{figure}

        We perform isolated training runs using individual datasets to enable direct comparisons between different datasets, such as Nemotron-CC-HQ(the high-quality subset of Nemotron-CC), CCI3-HQ(the high-quality subset of CCI3), and the whole CCI4.0 dataset introduced in this work. 
        The Figure \ref{fig:main} clearly demonstrates that across varying training scales (measured in tokens), the CCI4.0 dataset consistently outperforms both CCI3-HQ and Nemotron-CC-HQ, indicating superior training efficiency and better generalization performance. Key experimental findings are listed as follows:
\begin{itemize}
    \item At small-scale training ($\leq$20B tokens):
    CCI4.0 shows a significant performance advantage, particularly at 10B and 20B tokens, where it outpaces the other datasets by a notable margin. This suggests higher data quality and information density. For example, the performance of CCI4.0 at 10B tokens is comparable to Nemotron-CC-HQ at 30B tokens, highlighting its efficiency.
    \item At medium-scale training (20B–60B tokens):
    CCI4.0 continues to improve steadily and reaches performance saturation earlier than the other datasets, indicating that it enables the model to approach its performance ceiling more quickly under limited compute budgets.
    \item At large-scale training ($>$60B tokens):
    While the performance gap among datasets narrows, CCI4.0 maintains its lead, demonstrating that it remains effective even at larger training scales without exhibiting early overfitting or diminishing returns.
    \item Overall trend:
    CCI4.0 consistently delivers the highest scores across all training stages and exhibits a stable performance curve, indicating robust generalization and training stability.
\end{itemize}

\begin{table}[ht]
\centering
\caption{Evaluation results of different datasets across various benchmarks.}
\renewcommand{\arraystretch}{1.3} % 增加行高 (原始为1.0)
\setlength{\tabcolsep}{10pt}      % 增加列间距 (原始默认为6pt)
\resizebox{0.6\textwidth}{!}{
\begin{tabular}{@{}l S[table-format=2.2] S[table-format=2.2] S[table-format=2.2]@{}} % @{}移除了表格两端的额外空白
\toprule
\textbf{Metrics/Datasets} & {\textbf{CCI3-HQ}} & {\textbf{Nemotron-CC-HQ}} & {\textbf{CCI4.0}} \\ % S列的表头需要用{}包裹
\midrule
HellaSwag          & 28.06  & 44.63  & 42.50 \\
ARC (Average)      & 31.03  & 43.21  & 41.05 \\
PIQA               & 55.66  & 69.15  & 68.77 \\
MMLU (cloze)       & 26.52  & 30.32  & 30.34 \\
CommonsenseQA      & 21.05  & 27.19  & 27.44 \\
TriviaQA           & 1.28   & 5.91   & 6.05  \\
WinoGrande         & 48.62  & 51.38  & 51.46 \\
OpenBookQA         & 25.20  & 33.40  & 32.60 \\
SIQA               & 40.43  & 41.76  & 40.79 \\
CEval              & 32.31  & 27.74  & 27.67 \\
CMMLU              & 32.51  & 26.84  & 28.92 \\
\midrule % 使用 \midrule 代替 \hline 分隔数据和总结行
\textit{Average}$_{\textit{English}}$    & 30.87  & 38.55  & 37.89 \\
\textit{Average}$_{\textit{Chinese}}$   & 32.41  & 27.29  & 28.30 \\
\textit{Average}            & 31.64  & 32.92  & 33.09 \\
\bottomrule
\end{tabular}
}
\label{tab:evaluation-comparison-spacious}
\end{table}

The Table \ref{tab:evaluation-comparison-spacious} provides a comprehensive comparison of model performance across multiple benchmarks using three datasets: CCI3-HQ, Nemotron-CC-HQ, and CCI4.0. Although Nemotron-CC-HQ slightly outperforms CCI4.0 on the English average (38.55 vs. 37.89), CCI4.0 remains highly competitive and achieves the best score on several tasks such as CommonsenseQA (27.44) and TriviaQA (6.05). It also matches or closely trails Nemotron on others like MMLU (30.34 vs. 30.32), OpenBookQA (32.60 vs. 33.40), and SIQA (40.79 vs. 41.76), demonstrating robustness across diverse reasoning and knowledge benchmarks. One of CCI4.0’s most notable strengths lies in Chinese language tasks. It significantly outperforms Nemotron-CC-HQ in both CEval (27.67 vs. 27.74) and CMMLU (28.92 vs. 26.84), leading to a higher average Chinese score (28.30 vs. 27.29). This suggests that CCI4.0 contains a more representative and better-curated set of Chinese training data, leading to improved bilingual capabilities. Compared to CCI3-HQ, CCI4.0 shows major improvements in English benchmarks (especially ARC, MMLU, CommonsenseQA, and OpenBookQA), while exhibiting a slight disadvantage in Chinese performance. This is mainly due to the fact that Chinese data accounts for only around 20\% of the overall dataset composition. We aim for the CCI4.0 dataset to achieve a balanced trade-off between Chinese and English, making it a strong candidate for high-quality pretraining. 

% \begin{table}[ht]
% \centering
% \caption{Evaluation results of different datasets across various benchmarks.}
% \begin{tabular}{lccc}
% \hline
% \textbf{Metrics/Datasets} & \textbf{CCI3-HQ} & \textbf{Nemotron-CC-HQ} & \textbf{CCI4} \\
% \hline
% HellaSwag          & 28.06  & 44.63  & 42.50 \\
% ARC (Average)      & 31.03  & 43.21  & 41.05 \\
% PIQA               & 55.66  & 69.15  & 68.77 \\
% MMLU (cloze)       & 26.52  & 30.32  & 30.34 \\
% CommonsenseQA      & 21.05  & 27.19  & 27.44 \\
% TriviaQA           & 1.28   & 5.91   & 6.05  \\
% Winogrande         & 48.62  & 51.38  & 51.46 \\
% OpenBookQA         & 25.20  & 33.40  & 32.60 \\
% SIQA               & 40.43  & 41.76  & 40.79 \\
% CEval              & 32.31  & 27.74  & 27.67 \\
% CMMLU              & 32.51  & 26.84  & 28.92 \\
% \hline
% Average-English    & 30.87  & 38.55  & 37.89 \\
% Average-Chinese    & 32.41  & 27.29  & 28.30 \\
% Average            & 31.64  & 32.92  & 33.09 \\
% \hline
% \end{tabular}
% \label{tab:evaluation-comparison}
% \end{table}

\subsection{Reasoning Abilities from CoT Dataset}
\label{sec:reason}

To evaluate the impact of our CoT dataset on model reasoning abilities, we conducted a controlled experiment. Our evaluation approach is inspired by the adversarial CoT framework proposed by \cite{reflection2025}, which assesses model reflection on reasoning chains. Given the relatively small scale of our evaluated model, which does not exhibit emergent CoT generation capabilities, directly applying standard CoT evaluation methods is challenging. Therefore, we adapted the evaluation approach. For each test sample containing both a correct and an incorrect CoT, we measure the model's perplexity (PPL) on both CoTs. A sample is considered passed if the model assigns a lower PPL to the correct CoT compared to the incorrect one. The final score for a dataset is the proportion of samples that passed this PPL criterion. Following \cite{reflection2025}, we report the pre-training compute for each data point as $6nT$, where $n$ and $T$ are the number of parameters and training tokens, respectively.

\begin{figure}[H]
    \centering
    \includegraphics[width=1\textwidth]{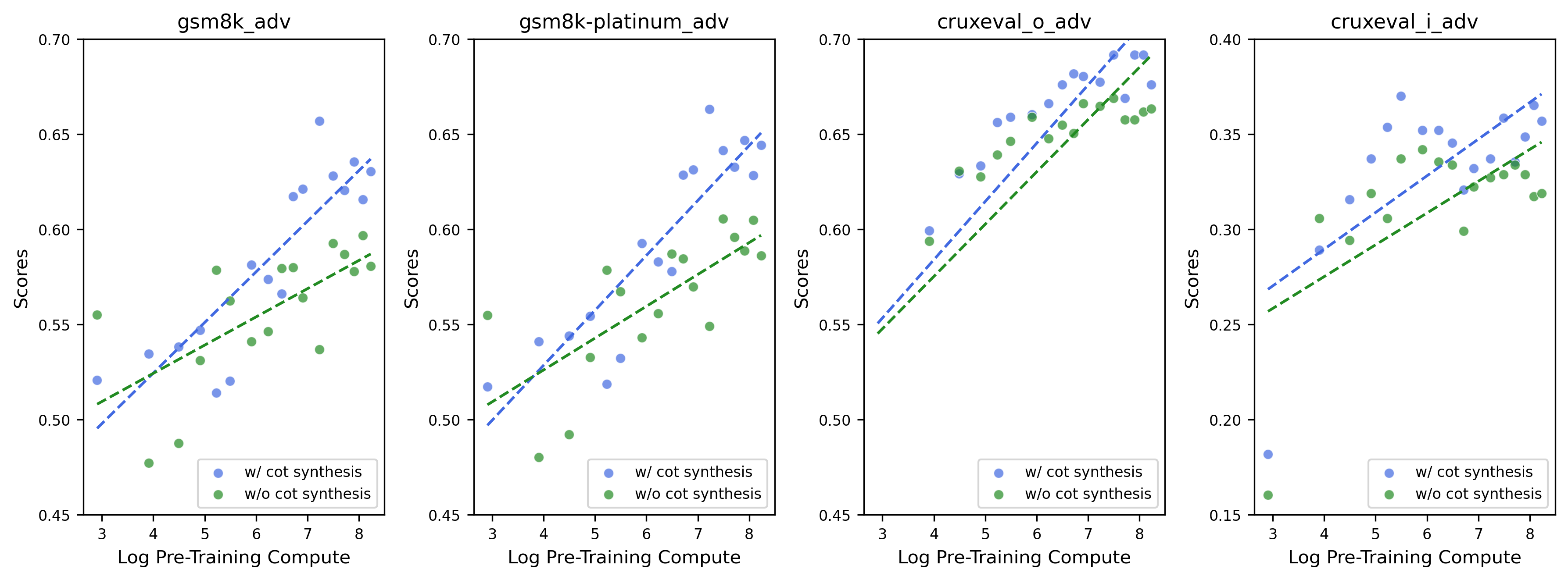}
    \caption{Reasoning ability scores on adversarial datasets for a 0.5B dense model trained with (w/ cot synthesis) and without (w/o cot synthesis) CoT data mix, evaluated across 100B training tokens. Training with CoT data accelerates reasoning ability growth.}
    \label{fig:reason1}
\end{figure}

We trained a 0.5B parameter dense language model on two distinct 100 billion-token datasets: one dataset included a mix of CoT data, while the other did not. We evaluated the reasoning performance of models checkpoints using our adapted PPL method on four adversarial datasets from gsm8k\_adv, gsm8k-platinum\_adv, cruxeval\_o\_adv, and cruxeval\_i\_adv. As shown in Figure \ref{fig:reason1}, compared to the model trained without CoT data, the model trained on the dataset mixed with CoT data demonstrates lower perplexity on correct CoT examples, indicating a notably faster improvement in reasoning ability across the evaluated datasets. This demonstrates that incorporating our CoT examples into the training data significantly reduces the model's tendency to hallucinate incorrect CoT examples, accelerating the acquisition of reasoning skills even in smaller models. Further experiments presented in Appendix \ref{app:reason} provide additional evidence that reasoning ability generally increases with the total training compute.

\subsection{Downstream Tasks Performance from CoT Dataset}

\begin{figure}[htbp] % H 表示将图片精确地放置在此处 (需要 float 宏包)
    % 其他可选的位置参数:
    % h: here - 尽量放在当前位置
    % t: top - 放在页面顶部
    % b: bottom - 放在页面底部
    % p: page of floats - 放在一个专门的浮动页面
    % !: 忽略 LaTeX 的一些内部参数，强制执行位置参数

    \centering % 图片居中显示
    \includegraphics[width=0.6\textwidth]{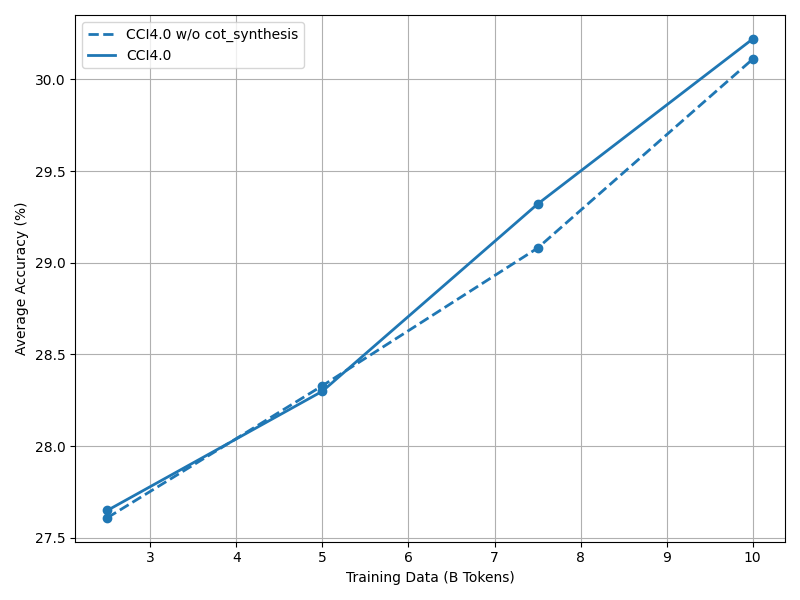} % 插入图片
    \caption{Performance Comparison of Models With and Without CoT Synthesis} % 图片的标
    \label{fig:avg_cot}
\end{figure}

To analyze the influence of CoT Datasets on the model performance, we provide the average model performance across downstream tasks in Figure \ref{fig:avg_cot}, where models are trained using 10 billion-token datasets with and without CoT data. More detailed model performance is provided in Table \ref{tab:cot_performance}. Results demonstrate that our synthetic CoT data contributes to performance gains in downstream tasks during model pretraining. Specifically, as shown in Table \ref{tab:cot_performance}, the model trained with CoT data performs well in reasoning tasks like HellaSwag and reading comprehension tasks like TriviaQA. However, the performance gains brought by CoT data to pretrained models on downstream reasoning tasks are inconsistent, and how to better leverage the effects of CoT data introduced during pretraining in the post-training stage warrants further investigation.

\section{Conclusion}
In this work, we introduced CCI4.0, a large-scale, bilingual pretraining dataset designed to provide high-quality and diverse coppora for model training. By integrating diverse, high-quality data sources, including Nemotron-CC for English and multiple Chinese datasets, alongside 4.5 billion human thinking templates via CCI4.0-M2-CoT, CCI4.0 addresses the limitations of traditional datasets in fostering general and complex reasoning abilities. The dataset’s rigorous processing pipeline—encompassing deduplication, multi-classifier quality scoring, fluency filtering, CoT synthesis, and privacy/toxicity handling—ensures both quality and diversity. Experimental results demonstrate that models pretrained on CCI4.0 significantly outperform baselines on reasoning-intensive benchmarks like MMLU and ARC-Challenge with a lower possibility of hallucination. These findings underscore the value of high-quality and diverse pretraining data and establish CCI4.0 as a new standard for developing LLMs capable of tackling sophisticated, multi-step reasoning challenges. Future work will explore further scaling and refinement of CoT synthesis to unlock even greater reasoning potential in next-generation models.

\section*{Acknowledgments}

We gratefully acknowledge the valuable contributions of Institutions Alibaba Cloud (阿里云), Shanghai AI Laboratory (上海人工智能实验室), Huawei (华为), Mobvoi (出门问问), Kingsoft Office Software (金山办公), Kunlun (昆仑万维), ModelBest (面壁智能), Qihoo (奇虎科技), Meituan (美团),  MiniMax (稀宇科技), Moonshot AI (月之暗面), Zidong Taichu (紫东太初), Wenge (中科闻歌) and iFLYTEK (科大讯飞) in providing the Chinese data.

% 阿里云、上海人工智能实验室、华为、出门问问、金山办公、昆仑万维、面壁智能、奇虎科技、美团、稀宇科技、月之暗面、紫东太初、中科闻歌、科大讯飞

\bibliography{neurips_2025}

\begin{thebibliography}{10}

\bibitem{reflection2025}
Essential AI, :, Darsh~J Shah, Peter Rushton, Somanshu Singla, Mohit Parmar,
  Kurt Smith, Yash Vanjani, Ashish Vaswani, Adarsh Chaluvaraju, Andrew Hojel,
  Andrew Ma, Anil Thomas, Anthony Polloreno, Ashish Tanwer, Burhan~Drak Sibai,
  Divya~S Mansingka, Divya Shivaprasad, Ishaan Shah, Karl Stratos, Khoi Nguyen,
  Michael Callahan, Michael Pust, Mrinal Iyer, Philip Monk, Platon Mazarakis,
  Ritvik Kapila, Saurabh Srivastava, and Tim Romanski.
\newblock Rethinking reflection in pre-training, 2025.

\bibitem{Bai2024COIGCQIAQI}
Yuelin Bai, Xinrun Du, Yiming Liang, Yonggang Jin, Ziqiang Liu, Junting Zhou,
  Tianyu Zheng, Xincheng Zhang, Nuo Ma, Zekun~Moore Wang, Ruibin Yuan, Haihong
  Wu, Hongquan Lin, Wenhao Huang, Jiajun Zhang, Wenhu Chen, Chenghua Lin, Jie
  Fu, Min Yang, Shiwen Ni, and Ge~Zhang.
\newblock Coig-cqia: Quality is all you need for chinese instruction
  fine-tuning.
\newblock {\em ArXiv}, abs/2403.18058, 2024.

\bibitem{banerjee-etal-2019-careful}
Pratyay Banerjee, Kuntal~Kumar Pal, Arindam Mitra, and Chitta Baral.
\newblock Careful selection of knowledge to solve open book question answering.
\newblock In Anna Korhonen, David Traum, and Llu{\'\i}s M{\`a}rquez, editors,
  {\em Proceedings of the 57th Annual Meeting of the Association for
  Computational Linguistics}, pages 6120--6129, Florence, Italy, July 2019.
  Association for Computational Linguistics.

\bibitem{baai_cci_data}
{Beijing Academy of Artificial Intelligence}.
\newblock {CCI-Data [Data set]}.
\newblock \url{https://huggingface.co/datasets/BAAI/CCI-Data}.

\bibitem{baai_cci2_data}
{Beijing Academy of Artificial Intelligence}.
\newblock {CCI2-Data [Data set]}.
\newblock \url{https://huggingface.co/datasets/BAAI/CCI2-Data}.

\bibitem{baai_wudaocorpora}
{Beijing Academy of Artificial Intelligence}.
\newblock {WuDaoCorporaText [Data set]}.
\newblock \url{https://data.baai.ac.cn/datadetail/WuDaoCorporaText}.

\bibitem{bisk2019piqareasoningphysicalcommonsense}
Yonatan Bisk, Rowan Zellers, Ronan~Le Bras, Jianfeng Gao, and Yejin Choi.
\newblock Piqa: Reasoning about physical commonsense in natural language, 2019.

\bibitem{Broder1997OnTR}
Andrei~Z. Broder.
\newblock On the resemblance and containment of documents.
\newblock {\em Proceedings. Compression and Complexity of SEQUENCES 1997 (Cat.
  No.97TB100171)}, pages 21--29, 1997.

\bibitem{Chen2023DataJuicer1}
Daoyuan Chen, Yilun Huang, Zhijian Ma, Hesen Chen, Xuchen Pan, Ce~Ge, Dawei
  Gao, Yuexiang Xie, Zhaoyang Liu, Jinyang Gao, Yaliang Li, Bolin Ding, and
  Jingren Zhou.
\newblock Data-juicer: A one-stop data processing system for large language
  models.
\newblock {\em Companion of the 2024 International Conference on Management of
  Data}, 2023.

\bibitem{chen2023chinesewebtext}
Jianghao Chen, Pu~Jian, Tengxiao Xi, Dongyi Yi, Qianlong Du, Chenglin Ding,
  Guibo Zhu, Chengqing Zong, Jinqiao Wang, and Jiajun Zhang.
\newblock Chinesewebtext: Large-scale high-quality chinese web text extracted
  with effective evaluation model, 2023.

\bibitem{clark2018thinksolvedquestionanswering}
Peter Clark, Isaac Cowhey, Oren Etzioni, Tushar Khot, Ashish Sabharwal, Carissa
  Schoenick, and Oyvind Tafjord.
\newblock Think you have solved question answering? try arc, the ai2 reasoning
  challenge, 2018.

\bibitem{Conneau2019UnsupervisedCR}
Alexis Conneau, Kartikay Khandelwal, Naman Goyal, Vishrav Chaudhary, Guillaume
  Wenzek, Francisco Guzm{\'a}n, Edouard Grave, Myle Ott, Luke Zettlemoyer, and
  Veselin Stoyanov.
\newblock Unsupervised cross-lingual representation learning at scale.
\newblock {\em ArXiv}, abs/1911.02116, 2019.

\bibitem{deepseekai2025deepseekv3technicalreport}
DeepSeek-AI, Aixin Liu, Bei Feng, Bing Xue, Bingxuan Wang, Bochao Wu, Chengda
  Lu, Chenggang Zhao, Chengqi Deng, Chenyu Zhang, Chong Ruan, Damai Dai, Daya
  Guo, Dejian Yang, Deli Chen, Dongjie Ji, Erhang Li, Fangyun Lin, Fucong Dai,
  Fuli Luo, Guangbo Hao, Guanting Chen, Guowei Li, H.~Zhang, Han Bao, Hanwei
  Xu, Haocheng Wang, Haowei Zhang, Honghui Ding, Huajian Xin, Huazuo Gao, Hui
  Li, Hui Qu, J.~L. Cai, Jian Liang, Jianzhong Guo, Jiaqi Ni, Jiashi Li, Jiawei
  Wang, Jin Chen, Jingchang Chen, Jingyang Yuan, Junjie Qiu, Junlong Li,
  Junxiao Song, Kai Dong, Kai Hu, Kaige Gao, Kang Guan, Kexin Huang, Kuai Yu,
  Lean Wang, Lecong Zhang, Lei Xu, Leyi Xia, Liang Zhao, Litong Wang, Liyue
  Zhang, Meng Li, Miaojun Wang, Mingchuan Zhang, Minghua Zhang, Minghui Tang,
  Mingming Li, Ning Tian, Panpan Huang, Peiyi Wang, Peng Zhang, Qiancheng Wang,
  Qihao Zhu, Qinyu Chen, Qiushi Du, R.~J. Chen, R.~L. Jin, Ruiqi Ge, Ruisong
  Zhang, Ruizhe Pan, Runji Wang, Runxin Xu, Ruoyu Zhang, Ruyi Chen, S.~S. Li,
  Shanghao Lu, Shangyan Zhou, Shanhuang Chen, Shaoqing Wu, Shengfeng Ye,
  Shengfeng Ye, Shirong Ma, Shiyu Wang, Shuang Zhou, Shuiping Yu, Shunfeng
  Zhou, Shuting Pan, T.~Wang, Tao Yun, Tian Pei, Tianyu Sun, W.~L. Xiao,
  Wangding Zeng, Wanjia Zhao, Wei An, Wen Liu, Wenfeng Liang, Wenjun Gao,
  Wenqin Yu, Wentao Zhang, X.~Q. Li, Xiangyue Jin, Xianzu Wang, Xiao Bi,
  Xiaodong Liu, Xiaohan Wang, Xiaojin Shen, Xiaokang Chen, Xiaokang Zhang,
  Xiaosha Chen, Xiaotao Nie, Xiaowen Sun, Xiaoxiang Wang, Xin Cheng, Xin Liu,
  Xin Xie, Xingchao Liu, Xingkai Yu, Xinnan Song, Xinxia Shan, Xinyi Zhou,
  Xinyu Yang, Xinyuan Li, Xuecheng Su, Xuheng Lin, Y.~K. Li, Y.~Q. Wang, Y.~X.
  Wei, Y.~X. Zhu, Yang Zhang, Yanhong Xu, Yanhong Xu, Yanping Huang, Yao Li,
  Yao Zhao, Yaofeng Sun, Yaohui Li, Yaohui Wang, Yi~Yu, Yi~Zheng, Yichao Zhang,
  Yifan Shi, Yiliang Xiong, Ying He, Ying Tang, Yishi Piao, Yisong Wang, Yixuan
  Tan, Yiyang Ma, Yiyuan Liu, Yongqiang Guo, Yu~Wu, Yuan Ou, Yuchen Zhu, Yuduan
  Wang, Yue Gong, Yuheng Zou, Yujia He, Yukun Zha, Yunfan Xiong, Yunxian Ma,
  Yuting Yan, Yuxiang Luo, Yuxiang You, Yuxuan Liu, Yuyang Zhou, Z.~F. Wu,
  Z.~Z. Ren, Zehui Ren, Zhangli Sha, Zhe Fu, Zhean Xu, Zhen Huang, Zhen Zhang,
  Zhenda Xie, Zhengyan Zhang, Zhewen Hao, Zhibin Gou, Zhicheng Ma, Zhigang Yan,
  Zhihong Shao, Zhipeng Xu, Zhiyu Wu, Zhongyu Zhang, Zhuoshu Li, Zihui Gu,
  Zijia Zhu, Zijun Liu, Zilin Li, Ziwei Xie, Ziyang Song, Ziyi Gao, and Zizheng
  Pan.
\newblock Deepseek-v3 technical report, 2025.

\bibitem{lighteval}
Clémentine Fourrier, Nathan Habib, Thomas Wolf, and Lewis Tunstall.
\newblock Lighteval: A lightweight framework for llm evaluation, 2023.

\bibitem{gandhi2025cognitive}
Kanishk Gandhi, Ayush Chakravarthy, Anikait Singh, Nathan Lile, and Noah~D
  Goodman.
\newblock Cognitive behaviors that enable self-improving reasoners, or, four
  habits of highly effective stars.
\newblock {\em arXiv preprint arXiv:2503.01307}, 2025.

\bibitem{gao2020pile}
Leo Gao, Stella Biderman, Sid Black, Chris Callison-Burch, Laurence Cohen, Esin
  Durmus, Ethan Fenoglio, Josh Firestone, Jordan Foster, Sam Gehman, Shachar
  Gretz, Kristen Hallahan, Dieuwke Hupkes, Nathan Lambert, Ron Le~Bras, Zachary
  Levonian, Luca Lisi, Annika McMillan-Major, Todor Mihaylov, Sewon Min, Colin
  Raffel, Melissa Roemmele, Baptiste Roziere, Maarten Sap, Vered Shwartz,
  Daniel Sileo, Sabari Subramanian, Leo Sutawika, Yoav Tewel, Josh Tow, Ethan
  Tseng, Luke van~der Poel, Vinay Venkatakrishnan, Benjamin Wang, Guillaume
  Wenzek, Thomas Wolf, Yuchen Wu, Yutong Xu, Xi~Yang, Michihiro Yasunaga, Peter
  Yin, Rowan Zellers, Tianwei Zhang, Jun Zhu, Mike Lewis, Federico Petroni, and
  Aleksandra Piktus.
\newblock The pile: An 800gb dataset of diverse text for language modeling.
\newblock {\em arXiv preprint arXiv:2101.00027}, 2020.

\bibitem{he2023wanjuancomprehensivemultimodaldataset}
Conghui He, Zhenjiang Jin, Chao Xu, Jiantao Qiu, Bin Wang, Wei Li, Hang Yan,
  Jiaqi Wang, and Dahua Lin.
\newblock Wanjuan: A comprehensive multimodal dataset for advancing english and
  chinese large models, 2023.

\bibitem{hendrycks2021measuringmassivemultitasklanguage}
Dan Hendrycks, Collin Burns, Steven Basart, Andy Zou, Mantas Mazeika, Dawn
  Song, and Jacob Steinhardt.
\newblock Measuring massive multitask language understanding, 2021.

\bibitem{huang2023ceval}
Yuzhen Huang, Yuzhuo Bai, Zhihao Zhu, Junlei Zhang, Jinghan Zhang, Tangjun Su,
  Junteng Liu, Chuancheng Lv, Yikai Zhang, Jiayi Lei, Yao Fu, Maosong Sun, and
  Junxian He.
\newblock C-eval: A multi-level multi-discipline chinese evaluation suite for
  foundation models.
\newblock In {\em Advances in Neural Information Processing Systems}, 2023.

\bibitem{joulin2016bag}
Armand Joulin, Edouard Grave, Piotr Bojanowski, and Tomas Mikolov.
\newblock Bag of tricks for efficient text classification.
\newblock {\em arXiv preprint arXiv:1607.01759}, 2016.

\bibitem{lee2022deduplicatingtrainingdatamakes}
Katherine Lee, Daphne Ippolito, Andrew Nystrom, Chiyuan Zhang, Douglas Eck,
  Chris Callison-Burch, and Nicholas Carlini.
\newblock Deduplicating training data makes language models better, 2022.

\bibitem{10.5555/3031843.3031909}
Hector~J. Levesque, Ernest Davis, and Leora Morgenstern.
\newblock The winograd schema challenge.
\newblock In {\em Proceedings of the Thirteenth International Conference on
  Principles of Knowledge Representation and Reasoning}, KR'12, page 552–561.
  AAAI Press, 2012.

\bibitem{li2024cmmlumeasuringmassivemultitask}
Haonan Li, Yixuan Zhang, Fajri Koto, Yifei Yang, Hai Zhao, Yeyun Gong, Nan
  Duan, and Timothy Baldwin.
\newblock Cmmlu: Measuring massive multitask language understanding in chinese,
  2024.

\bibitem{li2024datacomplmsearchgenerationtraining}
Jeffrey Li and Alex~Fang etc.
\newblock Datacomp-lm: In search of the next generation of training sets for
  language models, 2024.

\bibitem{OpenHermes2.5-zh}
Wenbo Pan.
\newblock Openhermes 2.5-zh: A partial chinese translation of openhermes-2.5,
  2024.

\bibitem{penedo2024fineweb-2}
Guilherme Penedo, Hynek Kydlíček, Vinko Sabolčec, Bettina Messmer, Negar
  Foroutan, Martin Jaggi, Leandro von Werra, and Thomas Wolf.
\newblock Fineweb2: A sparkling update with 1000s of languages, December 2024.

\bibitem{penedo2023refinedweb}
Guilherme Penedo, Quentin Malartic, Daniel Hesslow, Ruxandra Cojocaru,
  Alessandro Cappelli, Hamza Alibe, Quentin Chanu, Baptiste Launay,
  Jean-Baptiste Dehaene, and Hugo Touvron.
\newblock The refinedweb dataset for falcon llm: Outperforming curated corpora
  with web data, and web data only.
\newblock {\em arXiv preprint arXiv:2306.01116}, 2023.

\bibitem{deduplicate-text-datasets}
Google Research.
\newblock {Deduplicate Text Datasets}.
\newblock \url{https://github.com/google-research/deduplicate-text-datasets},
  2021.
\newblock Accessed: 2025-05-16.

\bibitem{sap2019socialiqacommonsensereasoningsocial}
Maarten Sap, Hannah Rashkin, Derek Chen, Ronan LeBras, and Yejin Choi.
\newblock Socialiqa: Commonsense reasoning about social interactions, 2019.

\bibitem{soldaini2024dolma}
Luca Soldaini, Rodney~Kin Lo, Wajdi Yazdan, Ahmed El-Kishky, Faisal Ladhak,
  Daniel Murray, Shaked Yom~Din, Winston Li, Yingbo Liu, Yanai Elazar, Akshita
  Bhagia, Dirk Groeneveld, Tim Dettmers, Aleksandra Piktus, Nicola Cancedda,
  Allie De~Lucia, Orr Katz, Leshem Choshen, Qiao Li, Hao Zhao, Avi Wettig,
  Alexander~M Rush, and Samuel~A Barnett.
\newblock Dolma: An open corpus of 3 trillion tokens for language model
  pretraining research.
\newblock {\em arXiv preprint arXiv:2402.00159}, 2024.

\bibitem{su2024nemotroncctransformingcommoncrawl}
Dan Su, Kezhi Kong, Ying Lin, Joseph Jennings, Brandon Norick, Markus Kliegl,
  Mostofa Patwary, Mohammad Shoeybi, and Bryan Catanzaro.
\newblock Nemotron-cc: Transforming common crawl into a refined long-horizon
  pretraining dataset, 2024.

\bibitem{su2024nemotron}
Dan Su, Kezhi Kong, Ying Lin, Joseph Jennings, Brandon Norick, Markus Kliegl,
  Mostofa Patwary, Mohammad Shoeybi, and Bryan Catanzaro.
\newblock Nemotron-cc: Transforming common crawl into a refined long-horizon
  pretraining dataset.
\newblock {\em arXiv preprint arXiv:2412.02595}, 2024.

\bibitem{huggingface2024fineweb}
HuggingFace Team.
\newblock The fineweb datasets: Decanting the web for the finest text data at
  scale.
\newblock {\em arXiv preprint arXiv:2406.17557}, 2024.

\bibitem{nvidia2024nemotron}
NVIDIA Team.
\newblock Nemotron-4 340b technical report.
\newblock {\em arXiv preprint arXiv:2412.02595}, 2024.

\bibitem{qwen2.5}
Qwen Team.
\newblock Qwen2.5: A party of foundation models, September 2024.

\bibitem{wang2024cci30hqlargescalechinesedataset}
Liangdong Wang, Bo-Wen Zhang, Chengwei Wu, Hanyu Zhao, Xiaofeng Shi, Shuhao Gu,
  Jijie Li, Quanyue Ma, TengFei Pan, and Guang Liu.
\newblock Cci3.0-hq: a large-scale chinese dataset of high quality designed for
  pre-training large language models, 2024.

\bibitem{qwen2}
An~Yang, Baosong Yang, Binyuan Hui, Bo~Zheng, Bowen Yu, Chang Zhou, Chengpeng
  Li, Chengyuan Li, Dayiheng Liu, Fei Huang, Guanting Dong, Haoran Wei, Huan
  Lin, Jialong Tang, Jialin Wang, Jian Yang, Jianhong Tu, Jianwei Zhang,
  Jianxin Ma, Jin Xu, Jingren Zhou, Jinze Bai, Jinzheng He, Junyang Lin, Kai
  Dang, Keming Lu, Keqin Chen, Kexin Yang, Mei Li, Mingfeng Xue, Na~Ni, Pei
  Zhang, Peng Wang, Ru~Peng, Rui Men, Ruize Gao, Runji Lin, Shijie Wang, Shuai
  Bai, Sinan Tan, Tianhang Zhu, Tianhao Li, Tianyu Liu, Wenbin Ge, Xiaodong
  Deng, Xiaohuan Zhou, Xingzhang Ren, Xinyu Zhang, Xipin Wei, Xuancheng Ren,
  Yang Fan, Yang Yao, Yichang Zhang, Yu~Wan, Yunfei Chu, Yuqiong Liu, Zeyu Cui,
  Zhenru Zhang, and Zhihao Fan.
\newblock Qwen2 technical report.
\newblock {\em arXiv preprint arXiv:2407.10671}, 2024.

\bibitem{yu2025opencsgchinesecorpusseries}
Yijiong Yu, Ziyun Dai, Zekun Wang, Wei Wang, Ran Chen, and Ji~Pei.
\newblock Opencsg chinese corpus: A series of high-quality chinese datasets for
  llm training, 2025.

\bibitem{yue2025does}
Yang Yue, Zhiqi Chen, Rui Lu, Andrew Zhao, Zhaokai Wang, Shiji Song, and Gao
  Huang.
\newblock Does reinforcement learning really incentivize reasoning capacity in
  llms beyond the base model?
\newblock {\em arXiv preprint arXiv:2504.13837}, 2025.

\bibitem{zellers2019hellaswagmachinereallyfinish}
Rowan Zellers, Ari Holtzman, Yonatan Bisk, Ali Farhadi, and Yejin Choi.
\newblock Hellaswag: Can a machine really finish your sentence?, 2019.

\end{thebibliography}
\newpage

\appendix

\section{Appendix}

% \begin{longtable}{>{\raggedright\arraybackslash}p{2.5cm}>{\raggedright\arraybackslash}p{5cm}>{\raggedright\arraybackslash}p{8cm}}
% \caption{Dataset Information} \label{tab:dataset_info} \\
% \toprule
% \textbf{Dataset Type} & \textbf{Dataset Name} & \textbf{URL} \\
% \midrule
\subsection{Loss values across domains}
\begin{figure}[H]
    \centering % 图片居中显示
    \includegraphics[width=1.0\textwidth]{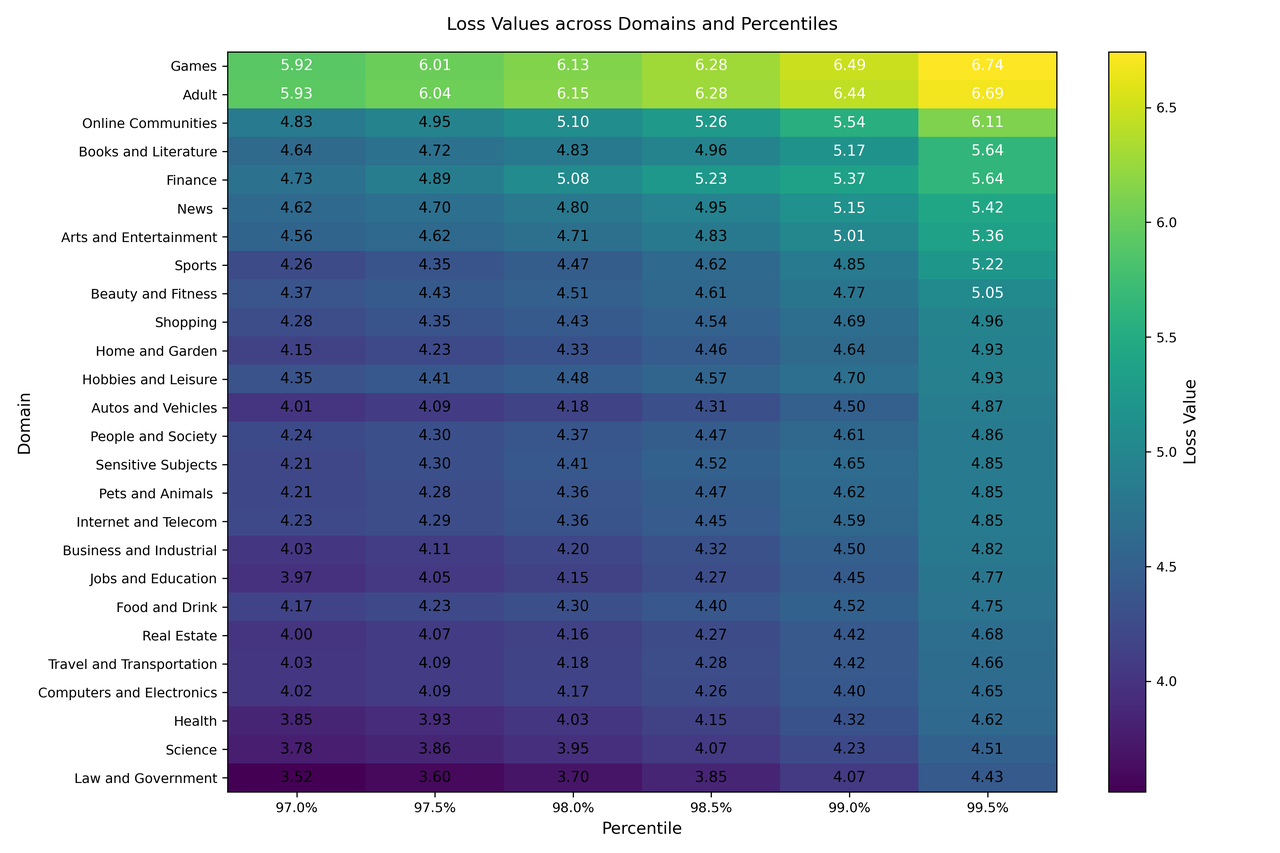} % 插入图片
    \caption{Loss values across domains and percentiles.} % 图片的标题
    \label{fig:loss}
\end{figure}
To systematically analyze the model's performance and guide our data filtering strategy, we investigated the distribution of loss values across various domains, as illustrated in Figure 5. The heatmap visualizes the loss values at high percentiles, ranging from the 97th to the 99.5th, offering a granular view of the most challenging instances within the dataset.

A key observation is the significant heterogeneity in loss distribution among the domains. Domains such as "Games" and "Adult" consistently exhibit higher loss values across all percentile thresholds, suggesting they contain a greater concentration of complex, noisy, or out-of-distribution samples that the model struggles to generalize. In contrast, domains like "Law and Government" and "Science" demonstrate substantially lower loss values, indicating a better model fit and cleaner data.

Based on this analysis, we established a data filtering threshold. The decision required a careful trade-off between removing potential noise and preserving valuable information, complicated but valid training examples. A lower percentile threshold would be too aggressive, potentially removing many informative samples. Therefore, we opted to perform the filtering at the 99.5th percentile. This conservative yet precise strategy targets only the most extreme outliers—the top 0.5\% of samples with the highest loss in each domain. This approach allows us to effectively prune the dataset of a majority of probable label errors and severe anomalies while retaining 99.5\% of the data, thus striking an optimal balance between enhancing data quality and maintaining the dataset's scale and diversity for robust model training.
\subsection{Data Sources} \label{dataset}
Table 3 provides a comprehensive list of the primary sources considered during our curation process. For the English component, we utilized $Nemotron-CC$, which is derived from Common Crawl. The Chinese component is more extensive, drawing from a variety of prominent web-scale datasets, including $WanJuan$, $WuDaoCorpora$, the $CCI-Data series$, and $fineweb-2$, among others. The inclusion of these varied and high-quality sources was crucial for ensuring the breadth, diversity, and scale necessary for our training objectives.

\renewcommand{\arraystretch}{1.3} % You can change 1.3 to other values like 1.5, 1.2 etc.

\begin{center} % Center the entire table
\begin{longtable}{>{\RaggedRight\arraybackslash}p{1.5cm}>{\RaggedRight\arraybackslash}p{6cm}>{\RaggedRight\arraybackslash}p{6cm}}
\caption{Main Datasets Considered During Dataset Curation} \label{tab:dataset_info} \\
\toprule
\textbf{Dataset Type} & \textbf{Dataset Name} & \textbf{URL} \\
\midrule
\endhead
Web-EN & Nemotron-CC & \url{https://data.commoncrawl.org/contrib/Nemotron/Nemotron-CC/index.html} \\
Web-ZH & WanJuan & \url{https://opendatalab.org.cn/OpenDataLab/WanJuan1_dot_0} \\
Web-ZH & CCI2.0-Data & \url{https://huggingface.co/datasets/BAAI/CCI2-Data} \\
Web-ZH & CCI1.0-Data & \url{https://huggingface.co/datasets/BAAI/CCI-Data} \\
Web-ZH & WudaoCourpora & \url{https://data.baai.ac.cn/details/WuDaoCorporaText} \\
Web-ZH & ChineseWebText2.0 & \url{https://huggingface.co/datasets/CASIA-LM/ChineseWebText2.0} \\
Web-ZH & MAP-CC & \url{https://huggingface.co/datasets/m-a-p/MAP-CC} \\
Web-ZH & TeleChat-PTD & \url{https://huggingface.co/datasets/Tele-AI/TeleChat-PTD} \\
Web-ZH & fineweb-2 & \url{https://huggingface.co/datasets/HuggingFaceFW/fineweb-2} \\
Web-ZH & HPLT Datasets v2 & \url{https://huggingface.co/datasets/HPLT/HPLT2.0_cleaned} \\
Web-ZH & CCI3.0-Data & \url{https://huggingface.co/datasets/BAAI/CCI3-Data} \\
Code & fineweb-code-corpus\_20241112 & \url{https://huggingface.co/datasets/OpenCoder-LLM/opc-fineweb-code-corpus} \\
Code & smollm-corpus-python-edu & \url{https://huggingface.co/datasets/HuggingFaceTB/smollm-corpus} \\
Math & fineweb-math-corpus\_20241112.jsonl & \url{https://huggingface.co/datasets/OpenCoder-LLM/opc-fineweb-math-corpus} \\
Math & EleutherAI-proof-pile-2-open-web-math.jsonl & \url{https://huggingface.co/datasets/EleutherAI/proof-pile-2} \\
Math & finemath-3plus.jsonl & \url{https://huggingface.co/datasets/HuggingFaceTB/finemath} \\
Books & dolma-books & \url{https://huggingface.co/datasets/allenai/dolma/blob/main/urls/v1_6.txt} \\
Wiki & dolma-wiki & \url{https://huggingface.co/datasets/allenai/dolma/blob/main/urls/v1_6.txt} \\
Arxiv & dolma-arxiv & \url{https://huggingface.co/datasets/allenai/dolma/blob/main/urls/v1_6.txt} \\
ForumQA & dolma-v1\_7-stackexchange & \url{https://huggingface.co/datasets/allenai/dolma/blob/main/urls/v1_7.txt} \\
\bottomrule

\end{longtable}
\end{center}

\subsection{Reasoning Ability Analysis}
\label{app:reason}
This section presents an experiment analyzing how reasoning ability scales with increasing training compute. We trained a 1.4B parameter MoE model (0.4B active) using 800 billion tokens of our proposed CoT data. We evaluated the model's reasoning performance using the adapted PPL method which is introduced in section \ref{sec:reason}.

\begin{figure}[htbp]
    \centering
    \includegraphics[width=1\textwidth]{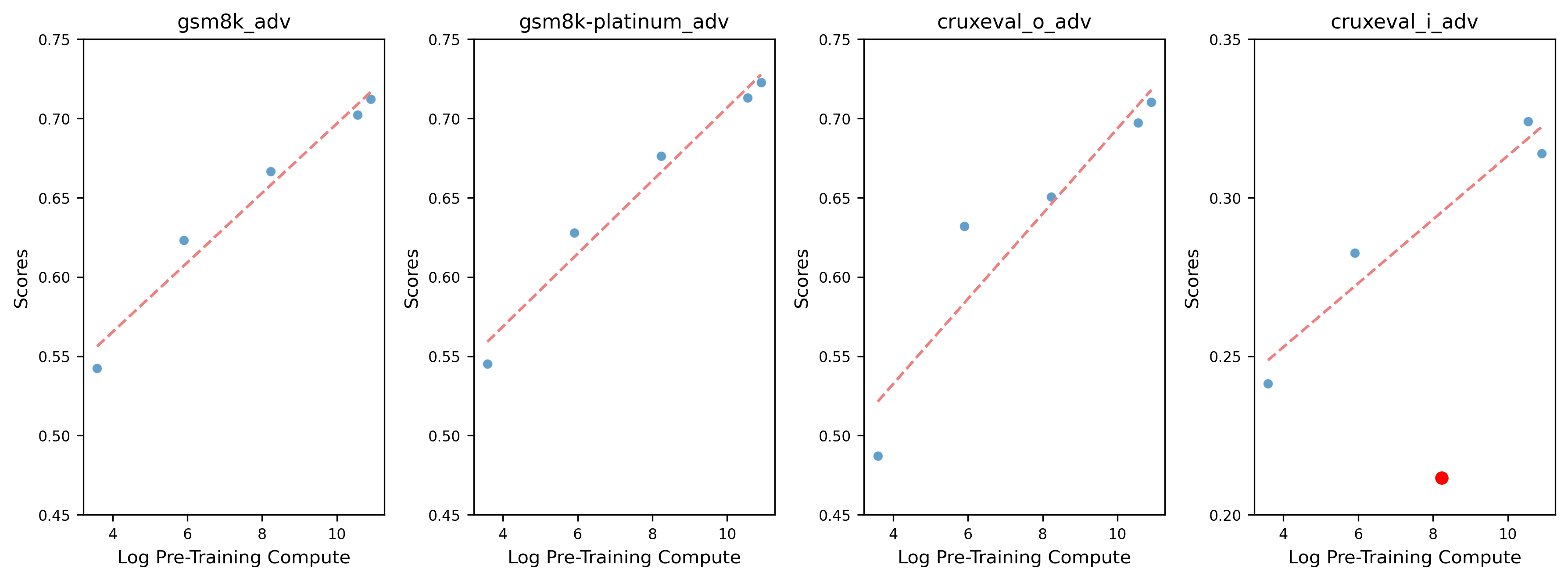}
    \caption{Reasoning ability scores on adversarial datasets for a 1.4B MoE model as pre-training compute increases (up to 800B tokens of CoT data). Reasoning ability improves consistently with increased training compute.}
    \label{fig:reason2}
\end{figure}

As illustrated in Figure \ref{fig:reason2}, the evaluation results demonstrate a clear trend: the model's reasoning ability shows a consistent improvement with increasing training compute on the CoT dataset. This finding suggests that training on a large volume of high-quality CoT data enhances the model's capacity to assign higher probability to correct reasoning paths, even in models where explicit CoT generation is not yet fully emergent.

\subsection{Downstream Task Performance}

We provide detailed performance of models trained with and without CoT synthesis dataset in Table \ref{tab:cot_performance}.
Results demonstrate that the model trained with CoT data performs well in reasoning tasks like HellaSwag and reading comprehension tasks like TriviaQA. However, the performance gains brought by CoT data to pretrained models on downstream reasoning tasks are inconsistent, and how to better activate the effects of these CoT data in the post-training stage warrants further investigation.

\begin{table}[htbp]
\centering
\caption{Detailed Performance of Models Trained With and Without CoT Synthesis.}
\renewcommand{\arraystretch}{1.3} % 增加行高
\setlength{\tabcolsep}{10pt}      % 增加列间距
\resizebox{0.85\textwidth}{!}{
\begin{tabular}{@{}l S[table-format=2.2] S[table-format=2.2]@{}}
\toprule
\textbf{Metrics/Datasets} & {\textbf{CCI4.0 without the CoT Data}} & {\textbf{CCI4.0}} \\
% \textbf{Metrics/Datasets} & {\textbf{Nemotron-CC-high}} & {\textbf{Nemotron-CC-high(from CCI4.0)}} \\
\midrule
HellaSwag          & 29.82 & 30.15 \\
ARC (Average)      & 33.03 & 33.16 \\
PIQA               & 62.79 & 61.26 \\
MMLU (cloze)       & 26.62 & 26.80 \\
CommonsenseQA      & 25.31 & 23.67 \\
TriviaQA           & 0.47  & 0.79  \\
WinoGrande         & 49.88 & 50.28 \\
OpenBookQA         & 28.00 & 28.20 \\
SIQA               & 40.99 & 40.43 \\
CEval              & 27.34 & 27.91 \\
CMMLU              & 27.11 & 27.46 \\
\midrule
\textit{Average}$_{\textit{English}}$    & 32.99 & 32.75 \\
\textit{Average}$_{\textit{Chinese}}$   & 27.23 & 27.69 \\
\textit{Average}                         & 30.11 & 30.22 \\
\bottomrule
\end{tabular}
}
\label{tab:cot_performance}
\end{table}

\subsection{Chinese Quality Classifiers}
\label{sec:quality}
We apply a combination of our three custom-built Chinese quality classifiers to categorize the Chinese dataset into different quality tiers. Similar to the approach used in Nemotron-CC, we validate the effectiveness of our classifiers by dividing the data into 20 buckets based on quality scores. Separate models are trained and evaluated using data from each bucket. For the Chinese dataset, we focus primarily on the average performance across Chinese evaluation benchmarks. 

\begin{figure}[htbp] % H 表示将图片精确地放置在此处 (需要 float 宏包)
    % 其他可选的位置参数:
    % h: here - 尽量放在当前位置
    % t: top - 放在页面顶部
    % b: bottom - 放在页面底部
    % p: page of floats - 放在一个专门的浮动页面
    % !: 忽略 LaTeX 的一些内部参数，强制执行位置参数
    \centering
    \includegraphics[width=0.7\textwidth]{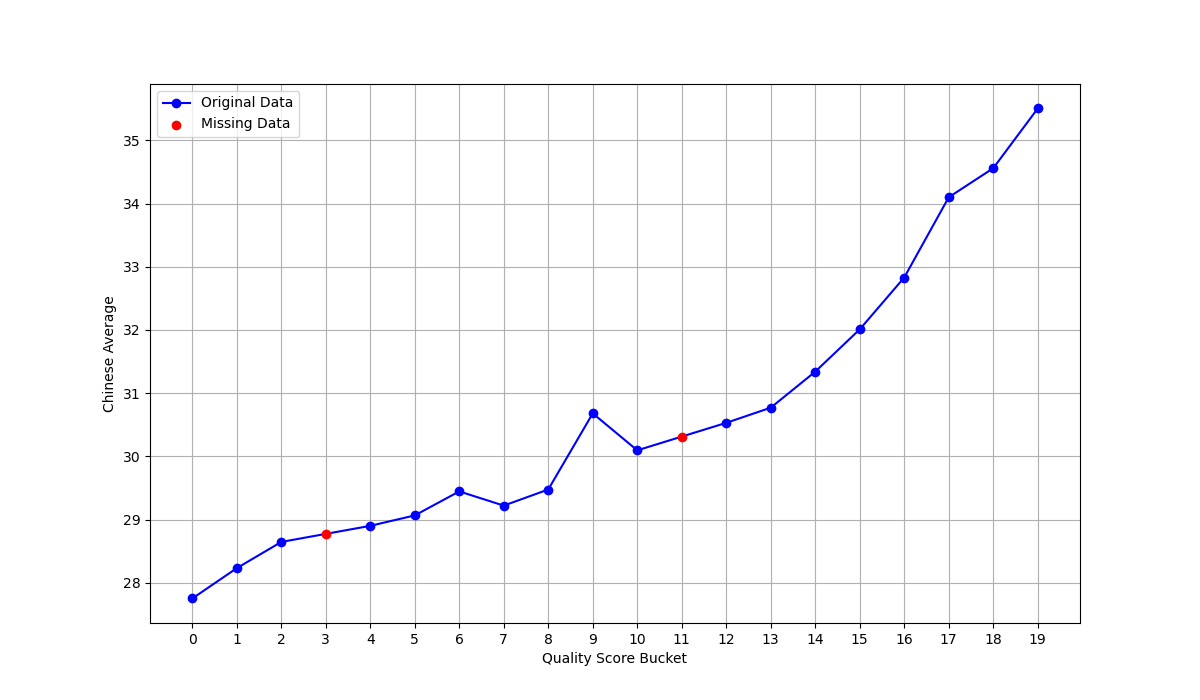} % 插入图片
    \caption{Performance Comparison of Different Datasets Across Training Scales. } % 图片的标
    \label{fig:clssubfig}
\end{figure}

The Figure~\ref{fig:clssubfig} presents the average Chinese evaluation scores of models trained on data buckets categorized by quality scores. Each bucket represents a range of quality scores as assigned by our Chinese data quality classifiers. Several key observations can be made: there is a clear upward trend in model performance as data quality increases. The average Chinese score improves steadily from bucket 0 to bucket 19, indicating that higher-quality data leads to significantly better downstream performance. This validates the effectiveness of the quality classifier in ranking data usefulness.

\subsection{Loss-Based Filtering}
\begin{figure}[htbp] % H 表示将图片精确地放置在此处 (需要 float 宏包)
    % 其他可选的位置参数:
    % h: here - 尽量放在当前位置
    % t: top - 放在页面顶部
    % b: bottom - 放在页面底部
    % p: page of floats - 放在一个专门的浮动页面
    % !: 忽略 LaTeX 的一些内部参数，强制执行位置参数

    \centering % 图片居中显示
    \includegraphics[width=0.65\textwidth]{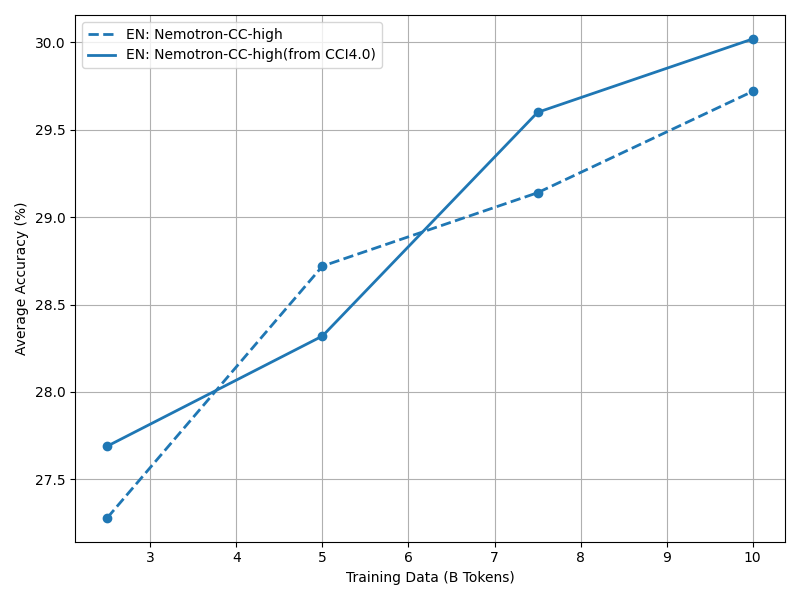} % 插入图片
    \caption{Performance Comparison of Models With and Without Loss-Based Filtering} % 图片的标
    \label{fig:losssubfig}
\end{figure}

To assess the effectiveness of loss-based filtering for English web data, we compare models trained on sampled 10 billion-token English corpora before and after loss filtering. Figure \ref{fig:losssubfig} presents the average performance during training. As shown in Figure~\ref{fig:losssubfig}, filtering based on loss improves training efficiency throughout the learning process. Table \ref{tab:loss_filter} further indicates that removing outlier samples with high loss from the raw English corpus enhances model performance on downstream commonsense reasoning tasks such as CommonsenseQA and SIQA, as well as reading comprehension tasks like TriviaQA. Notably, although this filtering method is applied solely to English web data, it also leads to slight performance improvements in Chinese QA tasks, such as those in the CMMLU benchmark.

\begin{table}[htbp]
\centering
\caption{Detailed Performance of Models Trained With and Without Loss-Based Filtering.}
\renewcommand{\arraystretch}{1.3} % 增加行高
\setlength{\tabcolsep}{10pt}      % 增加列间距
\resizebox{0.85\textwidth}{!}{
\begin{tabular}{@{}l S[table-format=2.2] S[table-format=2.2]@{}}
\toprule
\textbf{Metrics/Datasets} & {\textbf{Nemotron-CC-high}} & {\textbf{Nemotron-CC-high(from CCI4.0)}} \\
\midrule
HellaSwag          & 30.93 & 30.69 \\
ARC (Average)      & 34.39 & 34.43 \\
PIQA               & 62.73 & 62.89 \\
MMLU (cloze)       & 26.50 & 26.64 \\
CommonsenseQA      & 23.91 & 25.55 \\
TriviaQA           & 0.92  & 1.25  \\
WinoGrande         & 49.57 & 49.09 \\
OpenBookQA         & 29.60 & 30.20 \\
SIQA               & 39.56 & 40.12 \\
CEval              & 26.65 & 26.56 \\
CMMLU              & 25.98 & 26.67 \\
\midrule
\textit{Average}$_{\textit{English}}$    & 33.12 & 33.43 \\
\textit{Average}$_{\textit{Chinese}}$   & 26.32 & 26.62 \\
\textit{Average}                         & 29.72 & 30.02 \\
\bottomrule
\end{tabular}
}
\label{tab:loss_filter}
\end{table}

\subsection{Limitations} \label{limitation}
While our dataset offers broad coverage and high quality across Chinese and English data, it currently supports only these two languages. Future extensions will aim to incorporate additional languages to better support multilingual modeling and cross-lingual generalization.

Due to the large scale of the dataset, it may not be suitable for researchers with limited computational resources or those working with small models. In such cases, further filtering or subsetting of the dataset may be necessary to ensure practical usability.

Although we have applied privacy-preserving techniques and multiple toxicity filtering strategies using open-source models, we cannot guarantee the complete removal of all sensitive or harmful content. Users are advised to apply additional safeguards when deploying models trained on this dataset in sensitive applications.

\end{CJK}
\end{document}